\documentclass[runningheads]{llncs}
\usepackage{graphicx}
\usepackage{comment}
\usepackage{amsmath,amssymb} % define this before the line numbering.
\usepackage{color}
\usepackage{makecell}

\usepackage{xpatch}
\usepackage{multirow}

\usepackage[
  separate-uncertainty = true,
  multi-part-units = repeat
]{siunitx}

\usepackage[noend]{algpseudocode}
\DeclareMathOperator*{\argmax}{argmax}
\DeclareMathOperator*{\argmin}{argmin}

\begin{document}
% \linenumbers
\pagestyle{headings}
\mainmatter
\def\ECCVSubNumber{9}  % Insert your submission number here

\title{Active Class Incremental Learning for Imbalanced Datasets}% Replace with your title

%\begin{comment}
%\titlerunning{ECCV-20 submission ID %\ECCVSubNumber} 
%\authorrunning{ECCV-20 submission ID %\ECCVSubNumber} 
%\author{Anonymous ECCV submission}
%\institute{Paper ID \ECCVSubNumber}
%\end{comment}
%******************

%\begin{comment}
\titlerunning{Active Class Incremental Learning for Imbalanced Datasets}
%\author{Eden Belouadah\inst{1,2}%orcidID{0000-1111-2222-3333}
%\and
%Adrian Popescu\inst{1}%\orcidID{1111-2222-3333-4444} 
%\and
%Umang Aggarwal\inst{1}
%\and Léo Saci\inst{1}%\orcidID{2222--3333-4444-5555}
%}
\author{Eden Belouadah\inst{1,2}\orcidID{0000-0002-3418-1546}
\and Adrian Popescu\inst{1}\orcidID{0000-0002-8099-824X} 
\and Umang Aggarwal\inst{1}\orcidID{0000-0002-3982-9284} 
\and Léo Saci\inst{1}\orcidID{0000-0002-6957-7823}
}

\authorrunning{E. Belouadah et al.}
\institute{
%Université Paris-Saclay, CEA, Département Intelligence Ambiante et Systèmes Interactifs, 91191 Gif-sur-Yvette, France 
\small{
CEA, LIST, F-91191 Gif-sur-Yvette, France
\and
IMT Atlantique, Computer Science Department, F-29238, Brest, France\\
}
\email{\{eden.belouadah,adrian.popescu,umang.aggarwal,leo.saci\}@cea.fr}\\
}
%\end{comment}
%******************
\maketitle

\begin{abstract}
Incremental Learning (IL) allows AI systems to adapt to streamed data.
Most existing algorithms make two strong hypotheses which reduce the realism of the incremental scenario: (1) new data are assumed to be readily annotated when streamed and (2) tests are run with balanced datasets while most real-life datasets are imbalanced.
These hypotheses are discarded and the resulting challenges are tackled with a combination of active and imbalanced learning.
We introduce sample acquisition functions which tackle imbalance and are compatible with IL constraints.
We also consider IL as an imbalanced learning problem instead of the established usage of knowledge distillation against catastrophic forgetting.
Here, imbalance effects are reduced during inference through class prediction scaling.
Evaluation is done with four visual datasets and compares existing and proposed sample acquisition functions.
Results indicate that the proposed contributions have a positive effect and reduce the gap between active and standard IL performance.

\end{abstract}

%%%%%%%%% BODY TEXT
\section{Introduction}
AI systems are often deployed in dynamic settings where data are not all available at once~\cite{DBLP:journals/nn/ParisiKPKW19}.
Examples of applications include: (1) robotics - where the robot evolves in a changing environment and needs to adapt to it, (2) news analysis - where novel entities and events appear at a fast pace and should be processed swiftly, and (3) medical document processing - where parts of the data might not be available due to privacy constraints.

In such cases, incremental learning (IL) algorithms are needed to integrate new data while also preserving the knowledge learned for past data.
Following a general trend in machine learning, recent IL algorithms are all built around Deep Neural Networks (DNNs)~\cite{DBLP:conf/cvpr/AljundiCT17,Aljundi_2019_CVPR,DBLP:conf/eccv/CastroMGSA18,Hou_2019_CVPR,DBLP:conf/cvpr/MallyaL18,DBLP:conf/cvpr/RebuffiKSL17}.
The main challenge faced by such algorithms is catastrophic interference or forgetting~\cite{mccloskey:catastrophic}, a degradation of performance for previously learned information when a model is updated with new data.

IL algorithm design is an open research problem if computational complexity should remain bounded as new data are incorporated and/or if only a limited memory is available to store past data.
These two conditions are difficult to satisfy simultaneously and existing approaches address one of them in priority.
A first research direction allows for model complexity to grow as new data are added~\cite{DBLP:conf/cvpr/AljundiCT17,Aljundi_2019_CVPR,DBLP:conf/cvpr/MallyaL18,DBLP:journals/corr/RusuRDSKKPH16,DBLP:conf/cvpr/WangRH17}.
They focus on minimizing the number of parameters added for each incremental update and no memory of past data is allowed. 
Another research direction assumes that model complexity should be constant across incremental states and implements rehearsal over a bounded memory of past data to mitigate catastrophic forgetting~\cite{DBLP:conf/eccv/CastroMGSA18,DBLP:conf/bmvc/He0SC18,DBLP:journals/corr/abs-1807-02802,DBLP:conf/cvpr/RebuffiKSL17,DBLP:journals/corr/abs-1904-01769}.
Most existing IL algorithms assume that new data are readily labeled at the start of each incremental step. 
This assumption is a strong one since data labeling is a time consuming process, even with the availability of crowdsourcing platforms. 
Two notable exceptions are presented in~\cite{Aljundi_2019_CVPR} and~\cite{Pernici_2018_CVPR} where the authors introduce algorithms for self-supervised face recognition.
While interesting, these works are applicable only to a specific task and both exploit pretrained models to start the process. 
Also, a minimal degree of supervision is needed in order to associate a semantic meaning (i.e. person names) to the discovered identities.
A second hypothesis made in incremental learning is that datasets are balanced or nearly so.
In practice, imbalance occurs in a wide majority of real-life datasets but also in research datasets constructed in controlled conditions.
Public datasets such as ImageNet~\cite{DBLP:conf/cvpr/DengDSLL009}, Open Images~\cite{OpenImages2} or VGG-Face2~\cite{DBLP:conf/fgr/CaoSXPZ18} are all imbalanced. 
However, most research works related to ImageNet report results with the ILSVRC subset~\cite{DBLP:journals/ijcv/RussakovskyDSKS15} which is nearly balanced. 

These two hypotheses limit the practical usability of existing IL algorithms.
We replace them by two weaker assumptions to make the incremental learning scenario more realistic.
First, full supervision of newly streamed data is replaced by the possibility to annotate only a small subset of these data.
Second, no prior assumption is made regarding the balanced distribution of new data in classes.
We combine active and imbalanced learning methods to tackle the challenges related to the resulting IL scenario.

The main contribution of this work is to adapt sample acquisition process, which is the core component of active learning (AL) methods, to incremental learning over potentially imbalanced datasets.
A two phases procedure is devised to replace the classical acquisition process which uses a single acquisition function. 
A standard function is first applied to a subset of the active learning budget in order to learn an updated model which includes a suboptimal representation of new data.
In the second phase, a balancing-driven acquisition function is used to favor samples which might be associated to minority classes (i.e. those having a low number of associated samples).
The data distribution in classes is updated after each sample labeling to keep it up-to-date.  
Two balancing-driven acquisition functions which exploit the data distribution in the embedding space of the IL model are introduced here.
The first consists of a modification of the core-set algorithm~\cite{DBLP:conf/iclr/SenerS18} to restrain the search for new samples to data points which are closer to minority classes than to majority ones.
The second function prioritizes samples which are close to the poorest minority classes (i.e. those represented by the minimum number of samples) and far from any of the majority classes.
The balancing-driven acquisition phase is repeated several times and new samples are successively added to the training set in order to enable an iterative active learning process~\cite{Settles10activelearning}.

A secondary contribution is the introduction of a backbone training procedure which considers incremental learning with memory as an instance of imbalanced learning.
The widely used training with knowledge distillation ~\cite{DBLP:conf/eccv/CastroMGSA18,Hou_2019_CVPR,DBLP:journals/corr/abs-1807-02802,DBLP:conf/cvpr/RebuffiKSL17,DBLP:journals/corr/abs-1904-01769} is consequently replaced by a simpler procedure which aims to reduce the prediction bias towards majority classes during inference~\cite{DBLP:journals/nn/BudaMM18}.
Following the conclusions of this last work, initial predictions are rectified by using the prior class probabilities from the training set.

Four public datasets designed for different visual tasks are used for evaluation.
The proposed balancing-driven sample acquisition process is compared with a standard acquisition process and results indicate that it has a positive effect for imbalanced datasets. 

%-------------------------------------------------------------------------
\section{Related Works}
We discuss existing works from incremental, imbalanced and active learning areas and focus on those which are most closely related to our contribution.
Class incremental learning witnessed a regain of interest and all recent methods exploit DNNs.
One influential class of IL methods build on the adaptation of fine tuning and exploit increasingly sophisticated knowledge distillation techniques to counter catastrophic forgetting~\cite{mccloskey:catastrophic}. 
\textit{Learning without Forgetting} ($LwF$)~\cite{DBLP:conf/eccv/LiH16} introduced this trend and is an inspiration for a wide majority of further IL works. 
\textit{Incremental Classifier and Representation Learning} ($iCaRL$)~\cite{DBLP:conf/cvpr/RebuffiKSL17} is one such work which uses $LwF$ and also adds a bounded memory of the past to implement replay-based IL efficiently.
$iCaRL$ selects past class exemplars using a herding approach. 
The classification layer of the neural nets is replaced by a nearest-exemplars-mean, which adapts nearest-class-mean~\cite{DBLP:journals/pami/MensinkVPC13}, to counter class imbalance. 
\textit{End-to-end incremental learning}~\cite{DBLP:conf/eccv/CastroMGSA18} uses a distillation component which adheres to the original definition of $LwF$ from~\cite{DBLP:journals/corr/HintonVD15}. 
A balanced fine tuning step is added to counter imbalance. 
As a result, a consequent improvement over $iCaRL$ is reported.
\textit{Learning a Unified Classifier Incrementally via Rebalancing} ($LUCIR$)~\cite{Hou_2019_CVPR} tackles incremental learning problem by combining cosine normalization in the classification layer, a less-forget constraint based on distillation and an inter-class separation to improve comparability between past and new classes. 
\textit{Class Incremental Learning with Dual Memory} ($IL2M$)~\cite{belouadah2019il2m} and \textit{Bias Correction} ($BiC$)~\cite{DBLP:conf/cvpr/WuCWYLGF19} are recent approaches that add an extra layer to the network in order to remove the prediction bias towards new classes which are represented by more images than past classes.

Classical active learning is thoroughly reviewed in~\cite{Settles10activelearning}. 
A first group of approaches exploits informativeness to select items for uncertain regions in the classification space.
Uncertainty is often estimated with measures such as entropy~\cite{Shannon1948}, least confidence first~\cite{DBLP:conf/aaai/CulottaM05} or min margin among top predictions~\cite{DBLP:conf/icdm/SchefferDW01}.
Another group of approaches leverages sample representativeness computed in the geometric space defined by a feature extractor.
Information density~\cite{DBLP:conf/emnlp/SettlesC08} was an early implementation of such an approach.
\textit{Core-set}, which rely on the classical \textit{K-centers} algorithm to discover an optimal subset of the unlabeled dataset, was introduced in~\cite{DBLP:conf/iclr/SenerS18}.

Recent active learning works build on the use of deep learning. 
The labeling effort is examined in~\cite{DBLP:conf/iclr/HuLAR19} to progressively prune labels as labeling advances. 
An algorithm which learns a loss function specifically for AL was proposed in~\cite{Yoo_2019_CVPR}. 
While very interesting, such an approach is difficult to exploit in incremental learning since the main challenge here is to counter data imbalance between new and past classes or among new classes. 
Another line of works proposes to exploit multiple network states to improve the AL process. 
\textit{Monte Carlo Dropout}~\cite{DBLP:conf/icml/GalIG17} uses softmax prediction from a model with random dropout masks.
In~\cite{DBLP:conf/cvpr/BeluchGNK18}, an ensemble approach which combines multiple snapshots of the same training process is introduced.
These methods are not usable in our scenario because they increase the number of parameters due to the use of multiple models.
We retain the use of the same deep model through incremental states to provide embeddings  and propose a stronger role for them during the sample acquisition process. Recently,~\cite{Aggarwal_2020_WACV} proposed a method which focuses on single-stage AL for imbalanced datasets. They exploit a pretrained feature extractor and annotate the unlabeled samples so as to favor minority classes.

Ideally, incremental updates should be done in a fully unsupervised manner~\cite{Yang_2016_CVPR} in order to remove the need for manual labeling.
However, unsupervised algorithms are not mature enough to capture dataset semantics with the same degree of refinement and performance as their supervised or semi-supervised counterparts. 
Closest to our work are the self-supervision approaches designed for incremental face recognition~\cite{Aljundi_2019_CVPR,Pernici_2018_CVPR}. 
They are tightly related to unsupervised learning since no manual labeling is needed, except for naming the person.
Compared to self-supervision, our approach requires manual labeling for a part of new data and has a higher cost.
However, it can be applied to any class IL problem and not only to specific tasks such as face recognition as it is the case for~\cite{Aljundi_2019_CVPR,Pernici_2018_CVPR}.

A comprehensive review of imbalanced object-detection problems is provided in~\cite{DBLP:journals/corr/abs-1909-00169}. The authors group these problems in a taxonomy depending on their class imbalance, scale imbalance, spatial imbalance or objective  imbalance. The study shows the increasing interest of the computer vision community in the imbalanced problems for their usefulness in real life situations.

\begin{figure}[t]
	\begin{center}
\includegraphics[width=\textwidth]{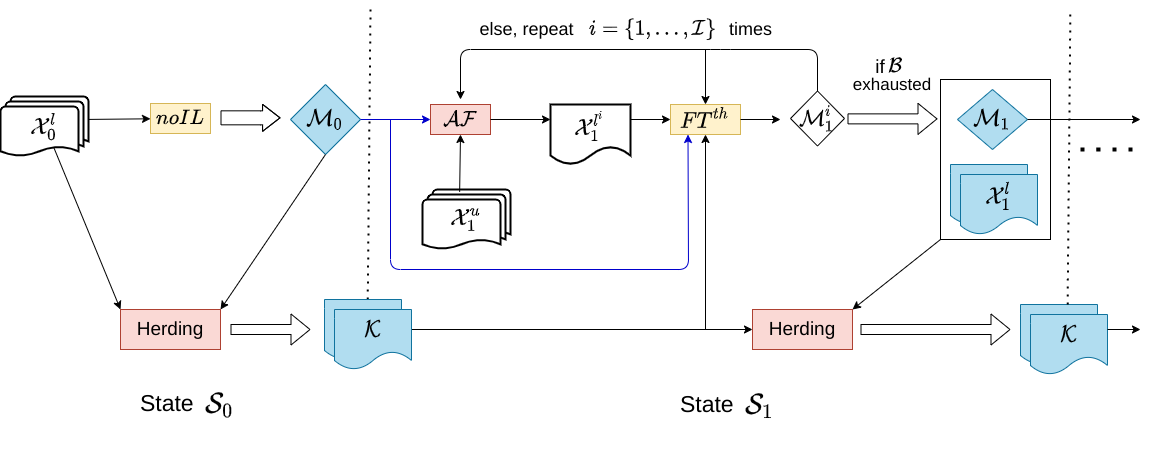}
	\end{center}
	
	%\vspace{-3em}
	
	\caption{Illustration of the proposed training process with one initial state $\mathcal{S}_0$, and one incremental state $\mathcal{S}_1$. The initial deep model $\mathcal{M}_0$ is trained from scratch on a fully-labeled dataset $\mathcal{X}_0^l$ using $noIL$ (a non-incremental learning). $\mathcal{M}_0$ and $\mathcal{X}_0^l$ are used to prepare the past class memory $\mathcal{K}$ using herding (a mechanism that selects the best representative past class images). State $\mathcal{S}_1$ starts with a sample acquisition function $\mathcal{AF}$ that takes the unlabeled set $\mathcal{X}_1^u$ and the model $\mathcal{M}_0$ as inputs, and provides a part of the budget $\mathcal{B}$ annotated as $\mathcal{X}_1^{l^i}$. The model $\mathcal{M}_0$ is then updated with data from $\mathcal{X}_{1}^{l^i}\cup \mathcal{K}$ using $FT^{th}$ (a fine tuning followed by a threshold calibration). The updated model $\mathcal{M}_1^i$ is again fed into the acquisition function $\mathcal{AF}$ with the rest of unlabeled examples from $\mathcal{X}_1^u$ to further annotate a part of the budget $\mathcal{B}$ and the model is updated afterwards. This process is repeated $\mathcal{I}$ times until $\mathcal{B}$ is exhausted. The model $\mathcal{M}_1$ is then returned with the annotated dataset $\mathcal{X}_1^l$ and the memory $\mathcal{K}$ is updated by inserting exemplars of new classes from $\mathcal{X}_1^l$ and reducing exemplars of past classes in order to fit its maximum size. Note that the two blue arrows are applicable in the first AL iteration only (when $i=1$). Best viewed in color.
	}
	\label{fig:overview}
\end{figure}

\section{Proposed Method}
The proposed active learning adaptation to an incremental scenario is motivated by the following observations:
\begin{itemize} 
\item Existing acquisition functions ($\mathcal{AF}$s) were designed and tested successfully for active learning over balanced datasets. However, a wide majority of real-life datasets are actually imbalanced. Here, no prior assumption is made regarding the imbalanced or balanced character of the unlabeled data which is streamed in IL states. Unlike existing sample acquisition approaches which exploit a single $\mathcal{AF}$, we propose to split the process in two phases. The first phase uses a classical $\mathcal{AF}$ to kick-off the process. The second one implements an  $\mathcal{AF}$ which is explicitly designed to target a balanced representation of labeled samples among classes. 
\item In IL, a single deep model can be stored throughout the process. This makes the application of recent ensemble methods~\cite{DBLP:conf/cvpr/BeluchGNK18} inapplicable. Following the usual AL pipeline, an iterative fine tuning of the model is implemented to incorporate labeled samples from the latest AL iteration.
\item A memory $\mathcal{K}$ of past class samples is allowed and, following~\cite{belouadah2019il2m,DBLP:conf/cvpr/HouPLWL19}, we model IL as an instance of imbalanced learning. The distillation component, which is central to most existing class IL algorithms~\cite{DBLP:conf/eccv/CastroMGSA18,DBLP:journals/corr/abs-1807-02802,DBLP:conf/cvpr/RebuffiKSL17,DBLP:conf/cvpr/WuCWYLGF19}, is removed. Instead, we deal with imbalance by using a simple but efficient post-processing step which modifies class predictions based on their prior probabilities in the training set. The choice of this method is motivated by its superiority in deep imbalanced learning over a large array of other methods~\cite{DBLP:journals/nn/BudaMM18}.
\end{itemize}

An illustration of the proposed learning process is provided in Fig.~\ref{fig:overview}. In the next sections, we first formalize the proposed active incremental learning scenario. 
Then, we introduce the adapted sample acquisition process, with focus on the balancing-driven acquisition functions.
Finally, we present the incremental learning backbone which is inspired from imbalanced learning~\cite{DBLP:journals/nn/BudaMM18}. 

%\vspace{-0.5em}

\subsection{Problem Formalization}
The formalization of the problem is inspired by~\cite{DBLP:conf/eccv/CastroMGSA18,DBLP:conf/cvpr/RebuffiKSL17} for the incremental learning part and by~\cite{DBLP:conf/cvpr/BeluchGNK18} for the active learning part.
We note $\mathcal{T}$ the number of states (including the first non-incremental state), $\mathcal{K}$ - the bounded memory for past classes, $\mathcal{B}$ - the labeling budget available for active learning, $\mathcal{AF}$ - an acquisition function designed to optimize sample selection in active learning, $\mathcal{I}$ the number of iterations done during active learning,
$\mathcal{S}_t$ - an incremental state, $N_t$ - the total number of classes recognizable in $\mathcal{S}_t$, $\mathcal{X}^{u}_{t}$ - the unlabeled dataset associated to $\mathcal{S}_t$, $\mathcal{X}^l_t$ - a manually labeled subset of $\mathcal{X}^u_t$, 
$\mathcal{M}_t$ - the deep model learned in $\mathcal{S}_t$.
The initial state $\mathcal{S}_0$ includes a dataset $\mathcal{X}_0=\{X^1_0, X^2_0, ...,X^j_0,..., X_0^{P_0}\}$ with $N_0 = P_0$ classes. $X^j_t=\{x_1^j,x_2^j, ..., x_{n_j}^j\}$ is the set of $n_j$ training examples for the $j^{th}$ class, $p_t^j$ is its corresponding classification probability in the state $\mathcal{S}_t$. 

We assume that all the samples are labeled in the first state. 
An initial non-incremental model $\mathcal{M}_0:\mathcal{X}_0 \rightarrow \mathcal{C}_0$ is trained to recognize a set $\mathcal{C}_0$ containing $N_0$ classes using all their data from $\mathcal{X}_0$.
$P_t$ new classes need to be integrated in each incremental state $\mathcal{S}_t$, with $t>0$. Each IL step updates the previous model $\mathcal{M}_{t-1}$ into the current model $\mathcal{M}_t$ which recognizes $N_t = P_0 + P_1 + ... + P_t$ classes in the incremental state $\mathcal{S}_t$. 
Active learning is deployed using $\mathcal{AF} (\mathcal{X}^u_t$) to obtain $\mathcal{X}^l_t$, a labeled subset from $\mathcal{X}_t^u$.

$\mathcal{X}^l_t$ data of the $P_t$ new classes are available but only a bounded exemplar memory $\mathcal{K}$ for the $N_{t-1}$ past classes is allowed.
$\mathcal{M}_t$, the model associated to the state $\mathcal{S}_t$ is trained over the $\mathcal{X}^l_t \cup \mathcal{K}$  training dataset.
An iterative AL process is implemented to recognize a set of classes $\mathcal{C}_t = \{c_t^1, c_t^2,..., c_t^{N_{t-1}}, c_t^{N_{t-1}+1},..., c_t^{N_t}\}$

%\vspace{-0.5em}

\subsection{Active Learning in an Incremental Setting}
\label{sub:alil}
We discuss the two phases of the adapted active learning process below.
Classical sampling is followed by a phase which exploits the proposed balancing-driven acquisition functions.

\subsubsection{Classical Sample Acquisition Phase.}

At the start of each IL state $\mathcal{S}_t$, an unlabeled dataset $\mathcal{X}_t^u$ is streamed into the system and classical AL acquisition functions are deployed to label $\mathcal{X}_t^l$, a part of $\mathcal{X}_t^u$, for inclusion in the training set.
Due to IL constraints, the only model available at the beginning of $\mathcal{S}_t$ is $\mathcal{M}_{t-1}$, learned for past classes in the previous incremental step.
It is used to extract the embeddings needed to implement acquisition functions.
A number of acquisition functions were proposed to optimize the active learning process~\cite{Settles10activelearning}, with adaptations for deep learning in~\cite{DBLP:conf/cvpr/BeluchGNK18,DBLP:conf/icml/GalIG17,DBLP:conf/iclr/SenerS18,DBLP:conf/cvpr/ZhouSZGGL17}.
Based on their strong experimental performance~\cite{DBLP:conf/cvpr/BeluchGNK18,DBLP:conf/icdm/SchefferDW01,DBLP:conf/iclr/SenerS18,Settles10activelearning}, four $\mathcal{AF}$s are selected for the initial phase:

\begin{itemize}
	\item \textbf{core-set sampling}~\cite{DBLP:conf/iclr/SenerS18} ($core$ hereafter): whose objective is to extract a representative subset of the unlabeled dataset from the vectorial space defined by the deep embeddings. The method selects samples with:
	\begin{equation}
		x_{next} = \argmax_{x_u \in \mathcal{X}_{t}^u}\{ \underset{1\leq k \leq n}{\min}\Delta (e(x_u), e(x_k)) \}		
		\label{eq:core}
	\end{equation}
	where: $x_{next}$ is the next sample to label, $x_u$ is an unlabeled sample left, $x_k$ is one of the $n$ samples which were already labeled, $e()$ is the embedding extracted using $\mathcal{M}_{t-1}$ and $\Delta$ is the Euclidean distance between two embeddings. 

	\item \textbf{random sampling} ($rand$ hereafter) : a random selection of images for labeling. While basic, random selection remains a competitive baseline in active learning.
	
	\item \textbf{entropy sampling}~\cite{Settles10activelearning} ($ent$ hereafter): whose objective is to favor most uncertain samples as defined by the set of probabilities given by the model. 
\begin{equation}
   x_{next} =  \argmax_{x_u \in \mathcal{X}_{t}^u}\{-\sum_{j=1}^{J}(p^j_t *\log(p^j_t))\}
      %x_{next} =  \argmax_{x_u \in \mathcal{X}_{u}^{N_t}}\{-\sum_{c=0}^{C}(P(y=c|x_u,w) * \log(P(y=c|x_u,w)))\}
    \label{eq:ent}
\end{equation}

where $p^j_t$ is the prediction score of $x_u$ for the class $j$ and $J$ is the number of detected classes so far by AL. 
%with: $H()$~\cite{Shannon1948} - the entropy of each unlabeled sample $x_u$ as calculated over $\mathcal{M}^{t-1}$ predictions.

\item \textbf{margin sampling}~\cite{DBLP:conf/icdm/SchefferDW01} ($marg$ hereafter): selects the most uncertain samples based on their top-2 predictions of the model. 
\begin{equation}
  x_{next} = \argmax_{x_u \in \mathcal{X}_{t}^u}\{max(p_t^1, .., p_t^j,.., p_t^J) - max_2(p_t^1, .., p_t^j,.., p_t^J)\}
%x_{next} = \argmax_{x_u \in \mathcal{X}_{u}^{N_t}}\{max(P(x_u)) - max_2(P(x_u))\}
\label{eq:margin}
\end{equation}
where $max(\cdot)$ and $max_2(\cdot)$ provide the top-2 predicted probabilities for the sample $x_u$.
This $\mathcal{AF}$ favors samples that maximize the difference between their top two predictions.

\end{itemize}

%This acquisition phase is launched once at the beginning of each incremental state to get an initial labeled subset of the new dataset. This step is necessary to expand the scores space by including new classes, these scores are indispensable to perform the first balancing-driven AL iteration.
This acquisition phase is launched once at the beginning of each incremental state to get an initial labeled subset of the new data.
This step is necessary to include the samples for the new classes in the trained model and initiate the iterative AL process.

\subsubsection{Balancing-driven Sample Acquisition Phase.}
The second acquisition phase tries to label samples so as to tend toward a balanced distribution among new classes. 
The distribution of the number of samples per class is computed after each sample labeling to be kept up-to-date.
The average number of samples per class is used to divide classes into minority and majority ones. These two sets of classes are noted $\mathcal{C}_t^{mnr}$ and $\mathcal{C}_t^{maj}$  for incremental state $\mathcal{S}_t$.
Two functions are proposed to implement the balancing-driven acquisition:

\begin{itemize}

\item \textbf{balanced core-set sampling} ($b-core$ hereafter) is a modified version of $core$ presented in Equation~\ref{eq:core}. $b-core$ acts as a filter which keeps candidate samples for labeling only if they are closer to a minority class than to any majority class. We write the relative distance of an unlabeled image w.r.t. its closest minority and majority classes as:
	\begin{equation}
	\small{
			\Delta_{\frac{mnr}{maj}}(x_{u}) = \min_{c_t^{mnr} \in \mathcal{C}_t^{mnr}} \Delta(e(x_u), \mu(c_t^{mnr})) - \min_{c_t^{maj} \in \mathcal{C}_t^{maj}} \Delta(e(x_u), \mu(c_t^{maj}))
		}
		\label{eq:delta_mnr}
	\end{equation}
	where: $x_u$ is an unlabeled sample, $c_t^{mnr}$ and $c_t^{maj}$ are classes from the minority and majority sets $\mathcal{C}_t^{mnr}$ and $\mathcal{C}_t^{maj}$ respectively, $e(x_u)$ is the embedding of $x_u$ extracted from the latest deep model available, $\mu(c_t^{mnr})$ and $\mu(c_t^{maj})$ are the centroids of minority and majority classes $c_t^{mnr}$ and $c_t^{maj}$ computed over the embeddings of their labeled samples.

The next sample to label is chosen by using the core-set definition from Equation~\ref{eq:core} but after filtering remaining unlabeled samples with Equation~\ref{eq:delta_mnr}:	

\begin{equation}
		x_{next} = \argmax_{x_u \in \mathcal{X}_t^u~and~\Delta_{\frac{mnr}{maj}}(x_{u}) < 0}\{ \underset{1\leq k \leq n}{\min}\Delta (e(x_u), e(x_k)) \}	
		\label{eq:bcore}
\end{equation}

\item \textbf{poorest class first sampling} ($poor$) is an acquisition function which gives priority to the class represented by the minimum number of labeled samples associated to it at a given moment during active learning. 
If there are several such classes, one of them is selected randomly.
The method translates the hypothesis that samples which are close to a poor class and far from any majority class should be favored in order to achieve a more balanced distribution.
The next candidate for labeling is selected with:
\begin{equation}
    x_{next} = \argmin_{x_u \in \mathcal{X}_{t}^u} \{ \Delta(e(x_u),\mu(c_t^{poor})) - \min_{\forall c_t^{maj} \in \mathcal{C}_t^{maj}} \Delta(e(x_u), \mu(c_t^{maj})) \} 
    \label{eq:poor}
\end{equation}
where $c_t^{poor}$ is a minority class from $\mathcal{C}_t^{mnr}$ which has the lowest number of samples in the current labeled subset.

$poor$ is similar in spirit to $b-core$ but has a stronger drive towards balancing because an individual class with poorest representation is targeted instead of samples which are close to any minority class. 
\end{itemize}

In an iterative active learning scenario, the balancing-driven acquisition can be repeated several time until the AL budget $\mathcal{B}$ is exhausted. 
%After each iteration, the model is updated through fine tuning by exploiting the sampled subset available at that moment. 

\subsection{Imbalance-driven Incremental Learning}
The model update within each incremental state is inspired by a usual iterative AL approach~\cite{Settles10activelearning} which includes a classical acquisition phase at the beginning and several balancing-driven iterations. For a total of $\mathcal{I}$ active learning iterations in each state $\mathcal{S}_t$, intermediate models $\mathcal{M}^{1}_{t}$, ..., $\mathcal{M}^{i}_{t}$, ..., $\mathcal{M}^{\mathcal{I}-1}_{t}$ are created while annotating $\mathcal{X}_t^{l^1}$, ..., $\mathcal{X}_t^{l^i}$, ..., $\mathcal{X}_t^{l^{\mathcal{I}-1}}$ during the first $\mathcal{I}-1$ iterations before obtaining the final $\mathcal{M}_{t}$.
The number of iterations $\mathcal{I}$ and the size of each iteration can take different values. 
The choice of a particular setting is done empirically so as to: (1) have enough new samples in the initial iteration in order for the new classes to be trainable in $\mathcal{M}^{1}_{t}$, i.e. the model $\mathcal{M}_{t}$ in the first iteration, (2) have enough candidates left for the balancing-driven iterations and (3) do not repeat the fine tuning process too many times to keep the incremental update timely. 
$\mathcal{M}_{t-1}$ is used to extract embeddings if $core$ is used in the initial AL iteration. Note that while iterative training increases the level of forgetting in IL~\cite{belouadah2019il2m,scail2020}, it is needed in AL to update model representation while annotating the images~\cite{Settles10activelearning}.

%This model is then successively fine tuned after each active learning iteration until the whole AL budget is used. 

As we mentioned, we depart from the usual modeling of the IL problem~\cite{DBLP:conf/eccv/CastroMGSA18,DBLP:conf/cvpr/HouPLWL19,DBLP:conf/cvpr/RebuffiKSL17,DBLP:conf/cvpr/WuCWYLGF19} which exploits knowledge distillation to counter catastrophic forgetting.
Following the recent observation that a simpler fine tuning based approach gives interesting results~\cite{belouadah2019il2m}, we use an IL backbone inspired from imbalance learning results presented in~\cite{DBLP:journals/nn/BudaMM18}. 
This backbone is called fine tuning with thresholding ($FT^{th}$ below), also known as threshold moving or post scaling~\cite{DBLP:journals/nn/BudaMM18}.
Thresholding adjusts the decision threshold of the model. It consists in the addition of a calibration layer at the end of the model during inference to compensate the prediction bias in favor of majority classes. 
This layer rectifies the class prediction $p^j_t$ of the $j^{th}$ class in the state $\mathcal{S}_t$ as follows:

\begin{equation}
    {p^j_t}^\prime = p^j_t \times {\frac{|\mathcal{X}^l_t \cup \mathcal{K}|}{|X_t^j|}} 
    \label{eq:ft_th}
\end{equation}

\noindent where $|X_t^j|$ is the number of training examples for the $j^{th}$ class in the state $\mathcal{S}_t$ and $| \mathcal{X}^l_t \cup \mathcal{K} |$ is the total number of training examples in state $\mathcal{S}_t$. 

$FT^{th}$  boosts the scores of classes with a lower number of associated samples.
The method has the interesting property of dealing with imbalance in IL in a uniform manner. 
It does not matter whether imbalance comes from the distribution of newly streamed data or from the fact that only a bounded memory of past classes is available.
This stands in contrast with knowledge distillation which handles imbalance for past classes but not among new ones. 
$FT^{th}$ is competitive against state-of-the-art algorithms. 
In a classical (i.e. fully supervised) IL setting, it has 59.59 top-1 accuracy for Cifar100, compared to $iCaRL$~\cite{DBLP:conf/cvpr/RebuffiKSL17} (57.35)
and $LUCIR$~\cite{DBLP:conf/cvpr/HouPLWL19} (55.36).  
More results are provided in the next section.

\section{Experiments}

\subsection{Datasets}
Experiments are run with four public datasets, out of which three are imbalanced and one is balanced. We provide a brief description of the datasets below:
\begin{itemize}
    \item \textbf{ImageNet100} - dataset for fine grained object recognition consisting of a subset of 100 randomly selected leaf classes from ImageNet~\cite{DBLP:conf/cvpr/DengDSLL009} which have at least 50 training images and are not present in the ILSVRC subset~\cite{DBLP:journals/ijcv/RussakovskyDSKS15}. 
    \item \textbf{Faces100} - face recognition dataset consisting of a subset of randomly selected 100 identities from VGG-Face2~\cite{DBLP:conf/fgr/CaoSXPZ18} with at least 30 training images.
    \item \textbf{Food101} - dataset for fine-grained food recognition~\cite{bossard14}. Since the initial dataset is perfectly balanced, an imbalance induction procedure was applied by removing a variable number of training samples keeping at least 25 images per class. 
    \item \textbf{Cifar100} - dataset for object recognition used in its original version~\cite{Krizhevsky09learningmultiple} which is perfectly balanced.
\end{itemize}

The main statistics of the experimental datasets are provided in Table~\ref{tab:dataset}.
We provide the coefficient of variation  $cv = \frac{\sigma}{\mu}$, with $\sigma$ the standard deviation and $\mu$ the mean of the distribution of samples per class.
$cv$ provides information about the degree of imbalance associated to each dataset.
The larger this value is, the more imbalanced the dataset will be.

\begin{table}[ht!]
    \begin{center}
    \resizebox{0.7\textwidth}{!}
    {
    \begin{tabular}{|c|c|c|c|c|c|c|}
        \hline
         Dataset & Classes & Train & Test & Mean train ($\mu$) & Std train ($\sigma$) & $cv$  \\ \hline
         ImageNet100 & 100 & 50000  & 5K  & 500.0 & 376.17   & 0.7523\\ \hline 
         Faces100   & 100 &   23237  & 5K  & 232.37 & 167.68  & 0.7216\\ \hline
         Food101 & 101 & 22374  & 10K  & 223.74 & 177.66 & 0.7940\\ \hline
         %Food101 &  22374  & 10K  & 223.74 & 180.66 & 0.8074\\ \hline
         Cifar100 & 100& 50000  & 10K  & 500.0 & 0.0  & 0.0\\ \hline 

    \end{tabular}
    }
    \end{center}
    \caption{Dataset statistics. $cv$ is the coefficient of variation defined as $cv = \frac{\sigma}{\mu}$. }
    \label{tab:dataset}
\end{table}

%\vspace{-4em}

\subsection{Methodology}
%$\mathcal{K}$ size is varied to evaluate the robustness of the tested methods with memory availability. 
\subsubsection{Incremental Learning Setting.}
We run the experiments with $\mathcal{T}=10$ states for each dataset\footnote{The initial non-incremental state of Food101 includes 11 classes while the initial states for the other datasets include 10 classes each.}. 
This setting is classically used in class incremental learning~\cite{DBLP:conf/eccv/CastroMGSA18,DBLP:conf/cvpr/RebuffiKSL17}. A total of $\mathcal{K}$ images of past classes are kept at any time during incremental learning.
 $\mathcal{K}$ approximates $2\%$ of the full training sets. 
Memory sizes are thus $\mathcal{K}=1000$ for ImageNet100 and Cifar100, $\mathcal{K}=465$ for Faces100 and $\mathcal{K}=450$ for Food101. 
At the end of each incremental state, memory is updated by inserting exemplars of new classes and reducing exemplars of past classes in order to fit its maximum size. 
 Note that since $\mathcal{K}$ is constant and the number of past classes grows, the imbalance in favor of new classes grows for later incremental states and the problem becomes more challenging.
The exemplars are chosen using the herding mechanism introduced in~\cite{DBLP:conf/cvpr/RebuffiKSL17}. 
The herding procedure consists in choosing the set of images that approximates the best the real mean of the class.

\subsubsection{Active Learning Process.}
Three active learning budgets are tested covering $\mathcal{B}=\{20\%, 10\%, 5\%\}$ of the unlabeled dataset $\mathcal{X}^u_t$ streamed in state $\mathcal{S}_t$. 
These different values are used to get a comprehensive view of each configuration's behavior.
Active learning is implemented with a usual iterative approach~\cite{DBLP:conf/iclr/SenerS18,Settles10activelearning} including $\mathcal{I}=4$ iterations, 40\% of $\mathcal{B}$ are used for classical acquisition and three times 20\% of $\mathcal{B}$ for balancing-driven acquisition (values were experimentally chosen).
Classical and balancing-driven acquisition phases are independent of one another and we test all their combinations.
For completeness, we include results with a baseline in which both phases are implemented with random sampling.
Note that the proposed acquisition functions are non-deterministic and experiments are run five times for each configuration in order to have a robust estimation of its performance.
To improve comparability of configurations which use the same initial $\mathcal{AF}$, the same initial models are used for all subsequent balancing-driven $\mathcal{AF}$s.

\subsubsection{Training Details.}
The experimental setup is inspired by the one proposed in $iCaRL$~\cite{DBLP:conf/cvpr/RebuffiKSL17}. 
$FT^{th}$ is implemented in Pytorch~\cite{paszke2017automatic} using a ResNet-18 architecture~\cite{DBLP:conf/cvpr/HeZRS16} and an SGD optimizer. 
The first non-incremental state is run for 100 epochs with $batch~size=128$, $lr=0.1$, $momentum=0.9$, $weight~decay=0.0005$. The learning rate is divided by 10 when the error plateaus for 15 consecutive epochs.
Fine tuning is run for 80 epochs, 20 epochs for each active learning iteration with $batch~size=32$, $lr=0.1$, $momentum=0.9$, $weight~decay=0.0005$. The learning rate is initialized at the beginning of the AL process and then divided by 10 when the error plateaus for 10 consecutive epochs.

Training images are preprocessed using randomly resized $224\times224$ crops and horizontal flipping and are normalized afterwards.
While more advanced data augmentation is known to slightly improve performance~\cite{DBLP:conf/eccv/CastroMGSA18}, we did not apply other image transformations.
For Faces100, face cropping is done with MTCNN~\cite{DBLP:journals/spl/ZhangZLQ16} before further processing.

\subsubsection{Upper Bound Methods.}
In addition to the active learning configurations, we present results with:
\begin{itemize}
\item{$sIL$} - usual supervised incremental learning in which all samples are labeled (equivalent to $\mathcal{B}=100\%$).
\item{$noIL$} - classical non-incremental learning in which all samples are provided at once.
\end{itemize}
For comparability, $sIL$ and $noIL$ are both trained using threshold calibration.
$sIL$ is an incremental upper bound for active learning configurations since it is fully supervised.
$noIL$ is an upper bound for $sIL$ since all the data are labeled and available at once.
These upper bounds are useful insofar they provide information about the performance gap due to a partial labeling of streamed data.

\subsection{Results and Discussion}
\subsubsection*{$FT^{th}$ in supervised mode} - Instead of handling catastrophic forgetting~\cite{mccloskey:catastrophic} as previous works did~\cite{DBLP:conf/eccv/CastroMGSA18,Hou_2019_CVPR,DBLP:conf/nips/RebuffiBV17,DBLP:conf/cvpr/WuCWYLGF19}, we address IL with bounded past memory as an imbalanced learning problem.
We use threshold calibration~\cite{DBLP:journals/nn/BudaMM18} to rectify scores in order to give more chances to minority classes to be selected during inference. 
The comparison to recent IL methods in supervised mode from Table~\ref{tab:supervised_results} indicates that $FT^{th}$ is competitive.
It clearly outperforms $iCaRL$~\cite{DBLP:conf/cvpr/RebuffiKSL17} and $IL2M$~\cite{belouadah2019il2m} and is better than LUCIR~\cite{Hou_2019_CVPR} for three datasets out of four.
We also provide the results of vanilla fine tuning before threshold calibration to underline the usefulness of thresholding.
It has a positive effect for all four datasets, a finding which validates its usefulness in our scenario.

\begin{table*}[h!]
%\captionsetup{width=.9\textwidth}
\begin{center}
\begin{tabular}{|c| c|c|c|c|c|}
\hline
Dataset & $FT$ & $FT^{th}$~\cite{DBLP:journals/nn/BudaMM18} & $LUCIR$~\cite{DBLP:conf/cvpr/HouPLWL19} & $iCaRL$~\cite{DBLP:conf/cvpr/RebuffiKSL17} & $IL2M$~\cite{belouadah2019il2m} \\
\hline
\hline

Imagenet100 &  54.80 & \textbf{61.42} & 60.77 & 52.40 & 57.68\\
\hline
Faces100 &  69.11 & 73.26 & \textbf{78.44} & 60.48 & 70.33 \\
\hline 
%real-  food ft = 34.15 , ft_th = 41.47 en offline
%current- food ft = ordinaire , ft_ th = en runtime
%real- il2m 36.20
%current- il2m 32.20 to don't get over ift th
%res icarl with lucir code: 50.46 - 69.45 - 20.70 - 44.55 
Food101 &  30.21 & \textbf{34.79} & 25.70 & 21.99 & 32.20\\
\hline
Cifar100 & 50.98 & \textbf{59.59} & 55.36 & 57.35 & 54.24 \\
\hline

\hline
\end{tabular}

\end{center}
%\vspace{-1em}
	\caption{Top-1 average supervised IL accuracy (\%). Best results are in bold.}
%\vspace{-1em}
\label{tab:supervised_results}
\end{table*}

\begin{table*}[t]
\begin{center}
\resizebox{\textwidth}{!}{
\begin{tabular}{|c|c|c|c|c|c|c|c|c|c|c|c|c|c|c|c|}
\hline
\multirow{2}{*}{Dataset} & \multirow{2}{*}{$\mathcal{B}$} & \multicolumn{3}{c|}{$rand$} & \multicolumn{3}{c|}{$core$}  & \multicolumn{3}{c|}{$ent$} & \multicolumn{3}{c|}{$marg$} & \multirow{2}{*}{\rotatebox[origin=c]{90}{\small $sIL$}} & \multirow{2}{*}{\rotatebox[origin=c]{90}{\small $noIL$}} \\
\cline{3-14}
& & $rand$ & $poor$ & $b-core$ & $core$ & $poor$ & $b-core$ & $ent$ & $poor$ & $b-core$ & $marg$ & $poor$ & $b-core$ & & \\
\hline
\multirow{3}{*}{\rotatebox[origin=c]{55}{\small ImageNet100}}
    & $20\%$ & \makecell{57.48\\ $\pm$0.50} & \makecell{\textbf{58.65}\\ $\pm$0.23} & \makecell{58.08\\ $\pm$0.40} & \makecell{56.85 \\ $\pm$0.23} & \makecell{57.25 \\ $\pm$0.84} & \makecell{57.46 \\ $\pm$0.52} & \makecell{45.07\\ $\pm$0.58} & \makecell{56.53\\ $\pm$0.27} & \makecell{56.23\\ $\pm$0.22} & \makecell{54.13\\ $\pm$0.66} & \makecell{56.39\\ $\pm$0.53} & \makecell{56.26\\ $\pm$0.45} & \multirow{3}{*}{\rotatebox[origin=c]{90}{\hspace{-3em} 61.42}} & \multirow{3}{*}{\rotatebox[origin=c]{90}{\hspace{-3em} 72.48}} \\ \cline{2-14}
        & $10\%$ &\makecell{52.61\\ $\pm$0.45} & \makecell{\textbf{54.89}\\ $\pm$0.53} & \makecell{53.40\\ $\pm$0.26} & \makecell{52.09\\ $\pm$0.41} & \makecell{53.55\\ $\pm$1.21} & \makecell{52.22\\ $\pm$1.13} & \makecell{42.15\\ $\pm$0.43} & \makecell{51.91\\ $\pm$0.51} & \makecell{51.81\\ $\pm$0.76} & \makecell{46.26\\ $\pm$1.24} & \makecell{51.40\\ $\pm$0.89} & \makecell{50.61\\ $\pm$0.36} &  &  \\ \cline{2-14} 
        & $5\%$ & \makecell{47.72\\ $\pm$0.69} & \makecell{\textbf{48.7}1\\ $\pm$0.97} & \makecell{48.18 \\ $\pm$0.56} & \makecell{46.01\\ $\pm$0.51} & \makecell{47.39\\ $\pm$0.85} & \makecell{46.45 \\ $\pm$0.30} & \makecell{37.95\\ $\pm$0.61} & \makecell{45.10\\ $\pm$1.46} & \makecell{44.70\\ $\pm$1.02} & \makecell{36.74\\ $\pm$1.05} & \makecell{44.18\\ $\pm$1.09} & \makecell{43.93\\ $\pm$0.93} & &  \\ \cline{2-14} 

\hline
\hline
\multirow{3}{*}{\rotatebox[origin=c]{55}{\small Faces100}}  
    & $20\%$ & \makecell{65.91\\ $\pm$0.94} & \makecell{66.41\\ $\pm$0.10} & \makecell{\textbf{67.24}\\ $\pm$0.36} & \makecell{66.41\\ $\pm$0.66} & \makecell{66.46\\ $\pm$0.46} & \makecell{66.94\\ $\pm$1.37} & \makecell{48.62\\ $\pm$0.95} &\makecell{63.30\\ $\pm$0.33} & \makecell{64.99\\ $\pm$0.80} & \makecell{59.51\\ $\pm$1.17} &\makecell{62.85\\ $\pm$1.75} & \makecell{65.27\\ $\pm$0.53} &  \multirow{3}{*}{\rotatebox[origin=c]{90}{\hspace{-3em} 73.26}} & \multirow{3}{*}{\rotatebox[origin=c]{90}{\hspace{-3em} 93.62}} \\ \cline{2-14} 
    
    & $10\%$ & \makecell{58.40\\ $\pm$0.71} & \makecell{\textbf{59.13}\\ $\pm$1.66} & \makecell{58.92\\ $\pm$1.05} & \makecell{55.82\\ $\pm$4.70} & \makecell{58.76\\ $\pm$2.93} & \makecell{57.26\\ $\pm$2.90} & \makecell{42.12\\ $\pm$1.38} & \makecell{54.93\\ $\pm$1.07} & \makecell{55.45\\ $\pm$1.47} & \makecell{49.32\\ $\pm$1.40} & \makecell{54.82\\ $\pm$2.19} & \makecell{58.69\\ $\pm$1.52}&  &  \\ \cline{2-14} 
    
    & $5\%$ & \makecell{48.38\\ $\pm$1.27} & \makecell{50.09\\ $\pm$2.30} & \makecell{\textbf{50.12}\\ $\pm$0.96} & \makecell{48.74\\ $\pm$1.21} & \makecell{47.71\\ $\pm$1.28} & \makecell{50.04\\ $\pm$2.04} & \makecell{35.61\\ $\pm$0.51} & \makecell{45.79\\ $\pm$0.83} & \makecell{45.39\\ $\pm$1.13} & \makecell{38.37\\ $\pm$1.02} & \makecell{45.90\\ $\pm$2.31} & \makecell{45.64\\ $\pm$1.56}  &  &  \\ \cline{2-14}

\hline
\hline
\multirow{3}{*}{\rotatebox[origin=c]{55}{\small Food101}}   
    & $20\%$ & \makecell{28.67\\ $\pm$0.42} & \makecell{\textbf{28.89}\\ $\pm$0.43} & \makecell{28.60\\ $\pm$0.52} & \makecell{28.24\\ $\pm$0.42} & \makecell{27.88\\ $\pm$0.34} & \makecell{27.98\\ $\pm$0.46} & \makecell{23.72\\ $\pm$1.00} & \makecell{27.99\\ $\pm$0.41} & \makecell{27.51\\ $\pm$0.54} & \makecell{28.13\\ $\pm$0.59} & \makecell{28.18\\ $\pm$0.35} & \makecell{27.56\\ $\pm$0.24} &  \multirow{3}{*}{\rotatebox[origin=c]{90}{\hspace{-3em} 34.79}} & \multirow{3}{*}{\rotatebox[origin=c]{90}{\hspace{-3em} 62.53}} \\ \cline{2-14} 
    
    & $10\%$ & \makecell{24.12\\ $\pm$0.47} & \makecell{\textbf{24.17}\\ $\pm$0.56} & \makecell{24.07\\ $\pm$0.68} & \makecell{22.91\\ $\pm$0.63} & \makecell{23.46\\ $\pm$0.12} & \makecell{23.07\\ $\pm$0.31} & \makecell{19.41\\ $\pm$0.96} & \makecell{22.32\\ $\pm$0.73} & \makecell{22.25\\ $\pm$0.64} & \makecell{23.35\\ $\pm$0.64} & \makecell{22.68\\ $\pm$0.81} & \makecell{22.84\\ $\pm$0.52}&  &  \\ \cline{2-14} 
    
    & $5\%$ & \makecell{20.51\\ $\pm$0.61} & \makecell{19.10\\ $\pm$0.68} & \makecell{\textbf{20.63}\\ $\pm$0.46} & \makecell{19.22\\ $\pm$0.36} & \makecell{19.17\\ $\pm$0.58} & \makecell{18.79\\ $\pm$0.64} & \makecell{16.80\\ $\pm$0.75} & \makecell{18.66\\ $\pm$0.31} & \makecell{18.57\\ $\pm$0.47} & \makecell{18.62\\ $\pm$0.48} & \makecell{18.79\\ $\pm$0.75} & \makecell{18.41\\ $\pm$0.82}  &  &  \\ \cline{2-14} 
                            
\hline
\hline
\multirow{3}{*}{\rotatebox[origin=c]{55}{\small Cifar100}}   
    & $20\%$ & \makecell{\textbf{49.47}\\ $\pm$0.16} & \makecell{49.36\\ $\pm$0.33} & \makecell{48.46\\ $\pm$0.75} & \makecell{46.75\\ $\pm$0.40} & \makecell{46.87\\ $\pm$0.19} & \makecell{46.87\\ $\pm$0.36} &  \makecell{39.76\\ $\pm$1.30} & \makecell{46.66\\ $\pm$0.29} & \makecell{47.69\\ $\pm$0.23} & \makecell{46.07\\ $\pm$0.31} & \makecell{45.37\\ $\pm$0.39} & \makecell{46.68\\ $\pm$0.44} &  \multirow{3}{*}{\rotatebox[origin=c]{90}{\hspace{-3em} 59.59}} & \multirow{3}{*}{\rotatebox[origin=c]{90}{\hspace{-3em} 76.98}} \\ \cline{2-14} 
    
    & $10\%$ & \makecell{\textbf{45.49}\\ $\pm$0.61} & \makecell{45.23\\ $\pm$1.17} & \makecell{44.83\\ $\pm$0.29} & \makecell{41.76\\ $\pm$0.54} & \makecell{42.60\\ $\pm$0.77} & \makecell{42.04\\ $\pm$0.77} & \makecell{34.87\\ $\pm$0.66} & \makecell{42.64\\ $\pm$0.50} & \makecell{43.76\\ $\pm$0.55} & \makecell{39.92\\ $\pm$0.43} & \makecell{40.94\\ $\pm$0.31} & \makecell{41.82\\ $\pm$0.35}&  &  \\ \cline{2-14} 
    
    & $5\%$ & \makecell{\textbf{41.58}\\ $\pm$0.29} & \makecell{40.69\\ $\pm$0.23} & \makecell{39.69\\ $\pm$0.46} & \makecell{35.23\\ $\pm$0.64} & \makecell{37.70\\ $\pm$0.46} & \makecell{35.72\\ $\pm$0.67} & \makecell{31.74\\ $\pm$0.74} & \makecell{37.68\\ $\pm$0.34} & \makecell{38.02\\ $\pm$0.66} & \makecell{31.88\\ $\pm$0.58} & \makecell{35.96\\ $\pm$0.42} & \makecell{35.69\\ $\pm$0.64}&  &  \\ \cline{2-14} 
\hline
\end{tabular}

}
\end{center}
%\vspace{-1em}
	\caption{Top-1 average accuracy (\%). Following~\cite{DBLP:conf/eccv/CastroMGSA18}, accuracy is averaged only for incremental states (i.e. excluding the initial, non-incremental state). Results are averaged over 5 runs for all AL configurations. $sIL$ is the result obtained in a fully supervised IL scenario. $noIL$ is the non-incremental upper-bound performance obtained with all data available. \textit{Best results for each active learning configuration (row) are in bold}.
	}
	
	%\vspace{-2em}

\label{tab:big}
\end{table*}

%\vspace{-2.5em}

\subsubsection*{Active Learning} - The experimental results obtained with $FT^{th}$ for the proposed active incremental learning scenario are presented in Table~\ref{tab:big}.  The comparison of classical $\mathcal{AF}$s ($rand$ -$rand$ and $core-core$ in Table~\ref{tab:big}) indicates that random sampling outperforms the $core-set$ sampling in a majority of cases.
This result is at odds with the one reported in~\cite{DBLP:conf/iclr/SenerS18} but is in line with the findings of~\cite{DBLP:conf/cvpr/BeluchGNK18,DBLP:conf/icml/GalIG17} that random sampling in AL is a strong baseline and is actually better than the recent core-set method from~\cite{DBLP:journals/corr/RusuRDSKKPH16}. The authors of this last paper also report that random sampling has better performance for lower active learning budgets which are studied here. Consequently, improving over random sampling for imbalanced datasets is an interesting result.

The results from Table~\ref{tab:big} indicate that the balancing-driven acquisition phase is useful for all three imbalanced datasets and active learning budgets tested. 
The gains for ImageNet100 and Faces100 are usually between $1$ and $2$ points compared to the classical acquisition processes implemented here ($rand$ - $rand$ or $core$ - $core$).
The gains are low for Food101, the third imbalanced dataset tested. 
This is probably due to the fact that Food101 is a more difficult task, as shown by $sIL$.
More labeled samples per class would probably be needed for an efficient training. 

$poor$ strategy is better than $b-core$ for ImageNet100 while more mixed results are obtained for Faces100 and Food101 datasets.
Interestingly, the best results are always obtained on top of a $rand$ initial sampling, even when $core$-$core$ baseline is better than a $rand$-$rand$ one, as it is the case for Faces100 with $\mathcal{B}=20\%$ and $\mathcal{B}=5\%$.

When applied without balancing, $ent$ and $marg$ have poorer performance compared to that of $rand$ and $core$.  Balancing significantly improves results for both of uncertainty-based methods, but their overall performance still lags behind that of random followed by balancing.
This reinforces the finding that a random selection is a competitive acquisition function in our active incremental learning over imbalanced datasets scenario.

The performance drop between active learning configurations and fully supervised IL naturally grows as $\mathcal{B}$ is reduced.
The drop between $sIL$ and the best AL configuration is of 3, 6 and 5 points for $\mathcal{B}=20\%$ for ImageNet100, Faces100, and Food101 respectively. 
When the AL budget is reduced to only 5\% of new data, the corresponding performance losses go to 12.5, 23 and 14 points.
Even when as little as 5\% of the new data are annotated, suboptimal models are trainable and usable if the IL system needs to be operational quickly. 

While the focus is on imbalanced datasets, we also report results with Cifar100, a perfectly balanced dataset for completeness.
In this case, the balancing-driven sampling has a slightly negative effect when applied over $rand$ and a slightly positive effect over $core$.
It is however notable that $core$ lags consistently behind $rand$ for Cifar100.
The best strategy for all $\mathcal{B}$ sizes is $rand$-$rand$, with $rand$ - $poor$ being a close second best configuration.

The gap between active IL and supervised IL is still notable, especially for smaller AL budgets.
In practice, active IL is useful when the system needs to be operational quickly after new data are streamed but at the expense of suboptimal performance. 
If a longer delay is permitted, it is naturally preferable to annotate all new data before updating the incremental model. 
The gap is even higher between incremental and classical learning, even though $FT^{th}$ has competitive performance compared to existing IL algorithms.
Globally, our results provide further confirmation that the use of incremental learning vs. classical learning should be weighted depending on the time, memory and/or computation constraints associated to an AI system's operation. 

%We complement the performance analysis from Table~\ref{tab:big} with a presentation of the imbalance ratios of active IL configurations. Check the supplementary material for more details.

%\vspace{-1em}

\section{Conclusion}

%\vspace{-0.5em}

We proposed a more realistic incremental learning scenario which does not assume that streamed data are readily annotated and that they are evenly distributed among classes.
An adaptation of the active learning sampling process is proposed in order to obtain a more balanced labeled subset.
This adaptation has a positive effect for imbalanced datasets and a slightly negative effect for the balanced dataset evaluated here.
Both proposed acquisition functions improve results compared to a classical acquisition process.
Also interesting, experiments show that the random baseline outperforms the $core-set$ function. 
The strong performance of random sampling indicates that this method should be consistently used as a baseline for future works in active incremental learning. 
As a secondary contribution, we introduce $FT^{th}$, a IL backbone which provides competitive results when compared to state-of-the-art methods. 
The code is publicly available to facilitate reproducibility\footnote{\url{https://github.com/EdenBelouadah/class-incremental-learning/}}.

The proposed approach brings the IL scenario closer to practical needs.
It reduces the time needed for an IL system to become operational upon receiving new data. 
The obtained results are encouraging but further investigation is needed to reduce the gap between active and supervised IL.  
Future work aims to: (1) run experiments with semi-supervised learning methods to automatically expand the labeled dataset and improve overall performance. While appealing, not all semi-supervised methods prove efficient in practice~\cite{NIPS2018_7585} and their usefulness for imbalanced datasets needs to be studied. 
(2) complement the proposed balancing-driven acquisition functions with a component which pushes the sampling process towards a better coverage of the manifold of each modeled class. This could be done, for instance, by taking inspiration from the herding mechanism~\cite{DBLP:conf/cvpr/RebuffiKSL17} already used to select past exemplars.
(3) render the IL scenario even more realistic by testing incremental steps of variable size to account for the fact that data might arrive at variable pace and considering that newly streamed data might belong both to unseen and past classes.\\

\noindent
\textbf{Acknowledgements}. This publication was made possible by the use of the FactoryIA supercomputer, financially supported by the Ile-de-France Regional Council.

%\clearpage\mbox{}Page \thepage\ of the manuscript.
%\clearpage\mbox{}Page \thepage\ of the manuscript.

%This is the last page of the manuscript.
%\par\vfill\par
%Now we have reached the maximum size of the ECCV 2020 submission (excluding references).
%References should start immediately after the main text, but can continue on p.15 if needed.

%\clearpage
\bibliographystyle{splncs04}
\bibliography{main}
\end{document}